# BIDA: A Bi-level Interaction Decision-making Algorithm for Autonomous Vehicles in Dynamic Traffic Scenarios

Liyang Yu[1], Tianyi Wang[2], Junfeng Jiao[3], Fengwu Shan[4], Hongqing Chu[1†], Bingzhao Gao[1,5]

*Abstract*—In complex real-world traffic environments, autonomous vehicles (AVs) need to interact with other traffic participants while making real-time and safety-critical decisions accordingly. The unpredictability of human behaviors poses significant challenges, particularly in dynamic scenarios, such as multi-lane highways and unsignalized T-intersections. To address this gap, we design a bi-level interaction decision-making algorithm (BIDA) that integrates interactive Monte Carlo tree search (MCTS) with deep reinforcement learning (DRL), aiming to enhance interaction rationality, efficiency and safety of AVs in dynamic key traffic scenarios. Specifically, we adopt three types of DRL algorithms to construct a reliable value network and policy network, which guide the online deduction process of interactive MCTS by assisting in value update and node selection. Then, a dynamic trajectory planner and a trajectory tracking controller are designed and implemented in CARLA to ensure smooth execution of planned maneuvers. Experimental evaluations demonstrate that our BIDA not only enhances interactive deduction and reduces computational costs, but also outperforms other latest benchmarks, which exhibits superior safety, efficiency and interaction rationality under varying traffic conditions.

## I. INTRODUCTION

Real-world traffic environments introduce significant challenges due to the unpredictability of diverse traffic participants (i.e., vehicles and pedestrians) in dynamic traffic environments (i.e., highway and urban scenarios) [1]–[3]. With the development of autonomous vehicles (AVs), machine learning and deep learning (DL) are increasingly being applied in the field of intelligent transportation systems to ensure safe and effective decision-making [4], [5]. Among these techniques, reinforcement learning (RL) stands out as a powerful method that enables agents to learn decision-making and motion-planning strategies, striving to achieve specific goals through trial-and-error learning [6]–[8].

Recent research has applied deep neural networks (DNNs) as function approximates for traditional policy and value functions in RL, which is known as deep reinforcement learning (DRL) [9]. Ragheb and Mahmoud [10] compared the performance of value-based deep Q-learning network and policy-based deep deterministic policy gradient (DDPG) in CARLA platform. Schulman et al. [11] proposed the trust region policy optimization (TRPO) to prevent large deviations from each update of the policy gradient and later designed the proximal policy optimization (PPO) to achieve better sample complexity and ensure smaller deviations. Wu et al. [12] further improved the PPO by combining a recurrent neural network to generate intrinsic reward signals, and introduced an auxiliary critic network to avoid overestimation bias based on the actor–critic framework. Regarding policy regularization, Haarnoja et al. [13] proposed the soft actor-critic (SAC), adding an entropy term to the reward function to improve robustness and stability of training.

To reflect real-world traffic characteristics, decision-making process can be accurately modeled as a partially observable Markov decision process (POMDP), which captures uncertainties in state estimation, future evolution of the traffic scene, and interactive behaviors [14]–[16]. Monte Carlo tree search (MCTS) algorithm has been widely adopted to solve POMDP problems in autonomous driving tasks [17]. MCTS's intrinsic ability to balance exploration and exploitation makes it well-suited for addressing the complex, dynamic, and uncertain nature of real-world traffic scenarios [18]. Karimi et al. [19] leveraged MCTS and level-k game theory to predict neighboring vehicles' future behaviors in lane-changing and ramp-merging scenarios, and realize real-time trajectory-planning. Wen et al. [14] extended MCTS to encompass broader behavior planning, integrating a diverse array of driving actions to handle key traffic scenarios.

Despite the advancements of RL and MCTS, pure RL-based approaches remain computationally intensive and prone to short sightedness, while pure MCTS-based approaches are restricted by insufficient intelligence and poor generalization. To overcome these limitations, hybrid approaches integrating RL with MCTS may serve as a viable solution [20]. The MCTS-based Alpha-Go Zero algorithm [21] mitigated the effect of distant rewarding, enhancing long-term decision-making. Building on this, Hoel et al. [22] created a tactical decision-making agent combining MCTS for planning and RL for training, where a DNN biased the sampling towards the most relevant parts of the search tree, while the MCTS component reduced both the required number of training samples and aids in finding long temporal correlations. Mo et al. [23] suggested an RL-based overtaking strategy that incorporated an RL agent and an MCTS algorithm to minimize unsafe behaviors in highway scenarios.

†Corresponding author: Hongqing Chu

[1]Liyang Yu, Hongqing Chu and Bingzhao Gao are with the School of Automotive Studies, Tongji University, Shanghai 201804, China. Email: yuliyang@tongji.edu.cn, chuhongqing@tongji.edu.cn, gaobz@tongji.edu.cn.

[2]Tianyi Wang is with the Department of Mechanical Engineering & Materials Science, Yale University, New Haven, CT 06511, USA. Email: tianyi.wang.tw727@yale.edu.

[3]Junfeng Jiao is with the School of Architecture, University of Texas at Austin, Austin, TX 78712, USA. Email: jjiao@austin.utexas.edu.

[4]Fengwu Shan is with Jiangxi Jiangling Group New Energy Vehicle Co., Ltd., Nanchang, Jiangxi 330052, China. Email: shanfw@jmev.com.

[5]Bingzhao Gao is also a permanent member of the Shanghai Key Laboratory of Wearable Robotics and Human-Machine Interaction.

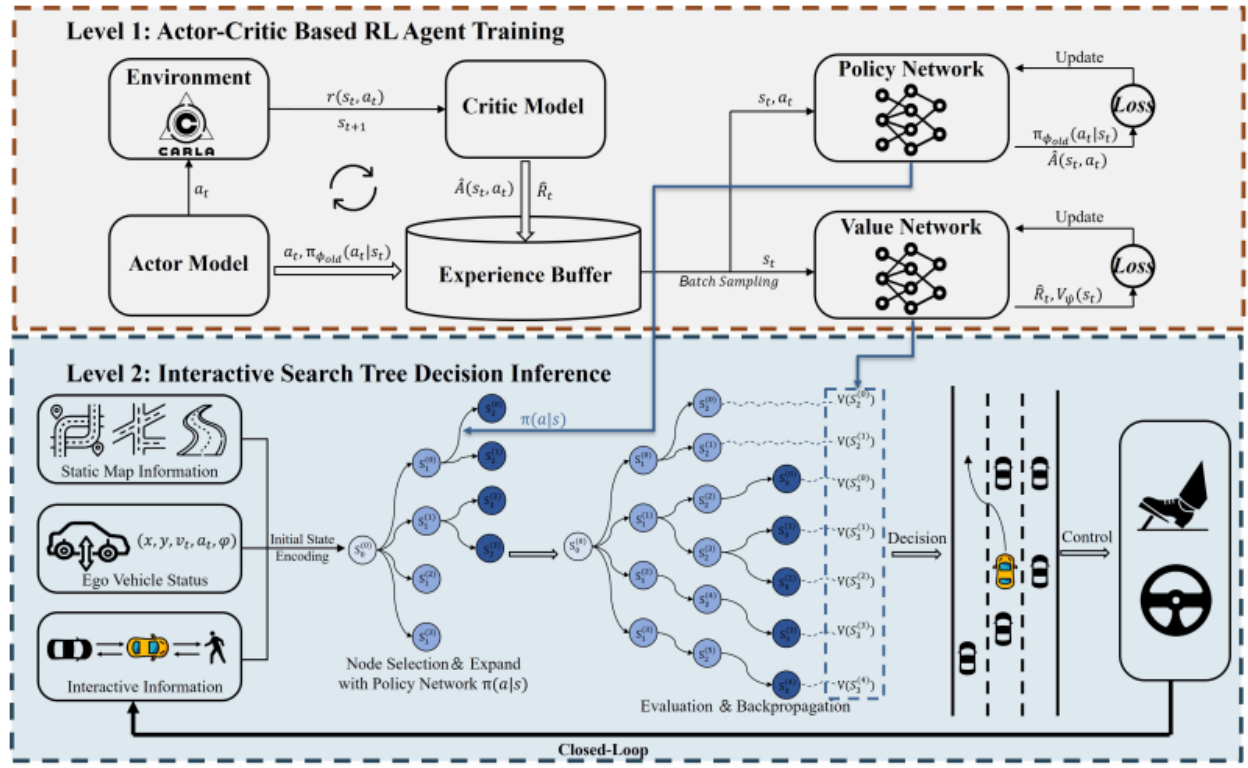

Fig. 1: Framework of the proposed bi-level interaction decision-making algorithm (BIDA): (1) *Upper Level*: DRL trains agents to obtain a reliable value network and policy network, which are applied to the online deduction process of interactive MCTS, assisting in value update and node selection. (2) *Lower Level*: After the optimal decisions are obtained, they are handed over to the planner and controller to execute the actions and a process of rolling optimization is achieved.

However, existing methods primarily utilized MCTS as a safety judgment mechanism in RL framework, failing to fully solve the problems of MCTS's computational costs and RL's short sightedness, i.e. lacking in-depth integration of MCTS and RL to improve the decision-making performance of AVs in dynamic traffic scenarios. To address these gaps, we introduce a bi-level interaction decision-making algorithm (BIDA) integrating MCTS with RL, as shown in Figure 1.

The main contributions of this paper are outlined as follows:

- **Bi-level Interaction Decision-making Algorithm:** *(1) Upper Level*: actor-critic-based RL agent training; *(2) Lower Level*: interactive search tree decision inference.
- **Rolling Optimization in Closed-loop Simulation:** *(1) A Closed-loop Simulation*: decision-making, motion-planning, and control in CARLA platform; *(2) A Rolling Optimization Process*: final decision handed over to the planner and controller.

## II. METHODOLOGY

### A. Actor-critic-based RL Agent Training

*1) Partially Observable Markov Decision Process:* Dynamic traffic scenarios involving AVs can be accurately modeled as a POMDP problem, which can be defined by Equation (1):

$$\mathcal{M} = \langle \mathcal{S}, \mathcal{A}, \mathcal{P}, \mathcal{R}, \mathcal{O}, \gamma \rangle, \quad (1)$$

where

- $\mathcal{S}$: The state space describing the environment.
- $\mathcal{A}$: The action space of the agent.
- $\mathcal{P} : \mathcal{S} \times \mathcal{A} \to \Delta(\mathcal{S})$: The state transition probability function, where $\Delta(\mathcal{S})$ denotes the probability distribution over $\mathcal{S}$.
- $\mathcal{R} : \mathcal{S} \times \mathcal{A} \to \mathbb{R}$: The reward function of the agent.
- $\mathcal{O}$: The observation space of the agent, which may only have partial observability of the environment.
- $\gamma \in [0, 1]$: The discount factor that weighs future rewards.

The agent aims to learn a policy $\pi : \mathcal{O} \to \mathcal{A}$ that maximizes its expected cumulative discounted reward:

$$\mathcal{J}(\pi) = \mathbb{E}\left[\sum_{t=0}^{\infty} \gamma^t \mathcal{R}(s_t, a_t)\right], \quad (2)$$

where subscript $t$ represents the time step.

*2) Reward Function:* The joint reward function is defined in Equation (3):

$$\mathcal{R} = w_{\text{success}} \cdot \mathcal{R}_{\text{success}} + w_{\text{safety}} \cdot \mathcal{R}_{\text{safety}} + w_{\text{comfort}} \cdot \mathcal{R}_{\text{comfort}}$$
$$+ w_{\text{efficiency}} \cdot \mathcal{R}_{\text{efficiency}} + w_{\text{interaction}} \cdot \mathcal{R}_{\text{interaction}} \quad (3)$$

where $w_{\text{success}}$, $w_{\text{safety}}$, $w_{\text{efficiency}}$, $w_{\text{comfort}}$, and $w_{\text{interaction}}$ are the weights regarding task success, traffic safety, traffic efficiency, ride comfort and interaction component of the reward functions, respectively; and $\mathcal{R}_{\text{success}}$, $\mathcal{R}_{\text{safety}}$, $\mathcal{R}_{\text{efficiency}}$, $\mathcal{R}_{\text{comfort}}$, and $\mathcal{R}_{\text{interaction}}$ are the corresponding reward functions that have been normalized.

For *task success*, it's worth noting that the agent's decision may conflict with the trajectory planner, and then the corresponding penalty should be given. $\mathcal{R}_{\text{success}}$ is calculated as:

$$\mathcal{R}_{\text{success}} = \begin{cases} -10 & \text{if collision happens,} \\ 10 & \text{if task is complete,} \\ -5 & \text{if no feasible solution exists.} \end{cases} \quad (4)$$

*Traffic safety* is paramount, which is captured by $\mathcal{R}_{\text{safety}}$:

$$\mathcal{R}_{\text{safety}} = -\frac{1}{1 + 0.2 d_{\min}^2} + \mathcal{N}_{\text{offroad}} \quad (5)$$

where $d_{\min}$ represents the minimum distance to other traffic participants; and $\mathcal{N}_{\text{offroad}}$ is the indicator to judge whether the vehicle drives off-road.

For *traffic efficiency*, $\mathcal{R}_{\text{efficiency}}$ rewards the maintenance of efficient traffic flow, which is calculated as:

$$\mathcal{R}_{\text{efficiency}} = \text{clip}(\frac{v_{ego}}{v_{target}}, 0, 1) \quad (6)$$

where $v_{ego}$ represents the ego vehicle's current velocity; and $v_{target}$ is the target velocity.

To ensure *ride comfort*, $\mathcal{R}_{\text{comfort}}$ initially penalizes sudden longitudinal acceleration and steering wheel angle changes:

$$\mathcal{R}_{\text{comfort}} = -0.5|a_x| - 0.2|\delta| \quad (7)$$

where $a_x$ and $\delta$ define the longitudinal acceleration change and steering wheel angle change, respectively.

The *initial interaction* component $\mathcal{R}_{\text{interaction}}$ evaluates how the agent interacts with other traffic participants:

$$\mathcal{R}_{\text{interaction}} = \sum_{i=0}^{N} a_i \text{ if } a_i < 0. \quad (8)$$

where $N$ represents the number of neighboring traffic participants; and $a_i$ is the deceleration, quantifying the negative impact caused by the agent's actions.

### B. Interactive Search Tree Decision Inference

Unlike the traditional MCTS algorithm, which approximates expected rewards through large-scale sampling, our BIDA estimates the Q-values using a value network and reward functions derived from DRL, which is computed as:

$$Q(s,a) = \mathbb{E}\left[\sum_t \gamma^t r_t\right] \quad (9)$$

where $r_t$ represents the reward at time step $t$.

The decision-making process of the interactive search tree is described in Algorithm 1. The details of the algorithm will be explained step-by-step as follows.

*1) Selection:* In the selection step, the AV selects an edge from a node based on a certain selection strategy, iteratively navigating the search tree until it reaches an unvisited edge. To improve decision-making efficiency, we incorporate an improved upper confidence bound (UCB) equation, which leverages a pre-trained RL policy network $\pi_\phi$:

$$\pi_\phi = \arg\max_a \left(Q(s,a) + c\pi_\phi(s,a)U(s,a)\right) \quad (10)$$

where the first term encourages the selection of actions with higher Q-values, while the second term promotes exploration, ensuring that actions not fully explored are still considered; $c$

**Algorithm 1** Interactive Search Tree Decision Inference
**Input:** Ego Vehicle Status, Static Map Information, Interactive Information
**Output:** Optimal Action
1: **for** $i = 1$ to $n$ **do**
2:    $t \leftarrow 0$
3:    $a_0 \leftarrow \arg\max_a (Q(s,a) + c\pi_\phi(s,a)U(s,a))$
4:    **while** $N(s_t, a_t) \neq 0$ **do** ▷ Selection
5:      $t \leftarrow t + 1$
6:      $a_t \leftarrow \arg\max_a (Q(s,a) + c\pi_\phi(s,a)U(s,a))$
7:    **end while**
8:    $s_{t+1} \leftarrow T_\theta(s_t, a_t)$ ▷ Expansion
9:    **for** each $a$ **do**
10:      $Q(s_{t+1}, a), N(s_{t+1}, a) \leftarrow 0, 0$
11:    **end for**
12:    **if** $s_{t+1}$ is terminal **then** ▷ Evaluation
13:      $R \leftarrow 0$
14:    **else**
15:      $R \leftarrow V_\psi(s_{t+1})$
16:    **end if**
17:    **for** $\tau = t, \ldots, 0$ **do** ▷ Backpropagation
18:      Set $(s, a, r) = (s_\tau, a_\tau, r_\tau)$
19:      $R \leftarrow r + \gamma R$
20:      $Q(s,a) \leftarrow Q(s,a) + \frac{1}{N(s,a)+1}(R - Q(s,a))$
21:      $N(s,a) \leftarrow N(s,a) + 1$
22:    **end for**
23: **end for**
24: $a^* \leftarrow \text{UCB\_Select}(Tree, s_0)$

is a constant; and $U(s,a)$ represents the degree of exploration at the node, which is defined as:

$$U(s,a) = \sqrt{\frac{\log(1 + \sum_a N(s,a))}{N(s,a) + 1}} \quad (11)$$

Besides, $\pi_\phi(s,a)$ is the output of the policy network, representing the probability of selecting action $a$ in the current state $s$, which satisfies:

$$\sum_{a \in A} \pi_\phi(s,a) = 1 \quad (12)$$

where $A$ is the same action space set as in DRL.

The policy network $\pi_\phi$ guides the search of tree nodes towards promising directions and the value network $V_\psi$ is used for evaluating tree nodes during the search process. This integration of DRL with MCTS allows for multi-step forward decision-making deduction, improving both computational efficiency and long-term planning in AV decision-making.

*2) Expansion:* During the expansion process, a new leaf node $s_{t+1}$ is generated from the selected edge $a_t$ in the selection step. In this paper, a predefined constant velocity model is used to achieve state transition, represented as:

$$s_{t+1} \leftarrow T_\theta(s_t, a_t) \quad (13)$$

In this model, it is assumed that neighboring vehicles and pedestrians maintain their current speeds in future actions,

thereby enabling the inference of future scenario states. The state transition model first samples the next state $s_{t+1}$, and the corresponding reward $r_t = r(s_t, a_t, s_{t+1})$ is computed using the reward function described in Section II-A.2. These values are then recorded in the new leaf node and used for the evaluation and backpropagation steps.

*3) Evaluation:* During the evaluation process, the pre-trained value network $V_\psi$ is used to evaluate the expected value of the newly expanded nodes. Given the value network $V_\psi$ and the reward $r$, the Q-value $Q(s, a)$ is estimated as:

$$Q(s, a) = r_t + \gamma V_\psi(s_{t+1}) \quad (14)$$

where $V_\psi(s_{t+1})$ denotes the expected reward.

Through the expansion step, the predicted state $s_{t+1}$ is obtained. Starting from this newly expanded node, the Q-value $Q(s, a)$ and visit count $N(s, a)$ for the connecting edges are initialized using the value network.

*4) Backpropagation:* In the backpropagation step, the values and rewards computed in the previous evaluation step are used to update all ancestor nodes of the newly expanded leaf node. The update process propagates from the leaf node to the root node, incrementally refining the visit count $N(s, a)$ and Q-value $Q(s, a)$ of each node. Each leaf node ancestor $s_\tau \in \{s_t, \ldots, s_0\}$ is updated according to the expected value $V_\psi(s_{t+1})$. The update rule follows:

$$Q(s_\tau, a_\tau) \leftarrow Q(s_\tau, a_\tau) + \frac{1}{N(s_\tau, a_\tau) + 1}(R_\tau - Q(s_\tau, a_\tau)) \quad (15)$$

where

$$R_\tau = \sum_{t'=\tau}^{t} \gamma^{t'-\tau} r_{t'} + \gamma^{t-\tau} V_\psi(s_{t+1}) \quad (16)$$

*5) Rolling Optimization:* To maintain real-time adaptability, a rolling optimization approach is adopted. The action associated with the edge from the root node to the highest-value child node is selected and executed as the optimal action. Then the environment's state changes due to the movement of other traffic participants. Consequently, the decision-making process is regenerated in the next step, continuously updating the search tree and planning process. This rolling optimization approach ensures that the behavior planning module can dynamically adapt to evolving traffic conditions, and make correct decisions in real-time.

## III. EXPERIMENTS

### A. Experimental Scenarios

The decision-making, motion-planning and control algorithms are fully implanted in CARLA [24], which ensures that the closed-loop simulation is conducted in a safe and controlled virtual environment. In this paper, two key traffic scenarios with high interaction characteristics are defined: a multi-lane highway scenario and an unsignalized T-intersection scenario. The schematic representation and key parameters of these two scenarios are shown in Figure 2 and Table I, respectively.

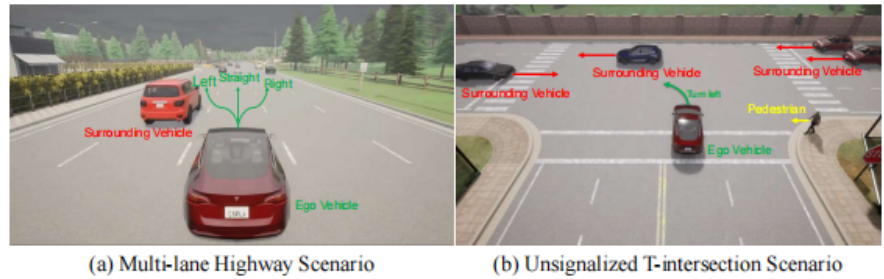

(a) Multi-lane Highway Scenario    (b) Unsignalized T-intersection Scenario

Fig. 2: Experimental traffic scenarios with high interactions

TABLE I: Parameters of the key traffic scenarios

| Parameters | Multi-lane Highway | Unsignalized T-intersection |
|---|---|---|
| Lane number | 5 | 2 |
| Lane width | 3.5 m | 3.5 m |
| Number of SVs* | 10/20/30 | 6/8/10 |
| Road speed limit | 120 km/h | 60 km/h |
| Simulation frequency | 20 Hz | 20 Hz |
| Speed zones for SVs | [80,120] km/h | [30,60] km/h |
| SV's behavior model | MOBIL | IDM |

*SV stands for Surrounding Vehicle.

*1) Multi-lane Highway:* This scenario simulates a suburban highway environment with five main lanes and a maximum speed limit of 120 km/h. The traffic flow is dynamic, incorporating a built-in intelligent driver model (IDM) that simulates realistic vehicle behaviors, including lane-changing, braking and reactions to reckless cut-in behaviors by the ego vehicle.

*2) Unsignalized T-intersection:* In this scenario, the ego vehicle needs to make a left turn from a side road onto the main road, which involves a complex traffic priority assessment and real-time decision-making process. We simulate the cross traffic flow, including pedestrians crossing the road, oncoming vehicles going straight on the main road, and vehicles turning left or right from the opposite direction.

*3) Closed-loop Experimental Settings:* In this paper, closed-loop simulation experiments are conducted in the CARLA simulator, where both the vehicle dynamics and environmental properties are preserved. The output decision commands generated by the proposed BIDA are first processed by a Lattice planner, which generates the reference trajectory. This trajectory is then tracked in real-time using a model predictive control (MPC) framework in a closed-loop manner. The optimization objective of MPC is to minimize trajectory tracking deviations, ensuring that the ego vehicle follows the planned path accurately while maintaining stability. The CasADi solver is used for numerical optimization. The vehicle kinematic model is employed to construct the controller's plant, cost function and constraints for MPC.

### B. Training Results

Given that the BIDA simultaneously relies on both a policy network and a value network, we adopt three actor-critic-based RL algorithms (i.e., TRPO, PPO and SAC) to train the ego vehicle agent for the two dynamic interaction scenarios described above. For both traffic scenarios, the TRPO, PPO and SAC models are trained for 100,000 update steps, with 10 repetitions using different random seeds. The training results are shown in Figure 3, where the dark solid line represents the mean of the results from the 10 experiments, and the shaded region represents the 90% confidence interval.

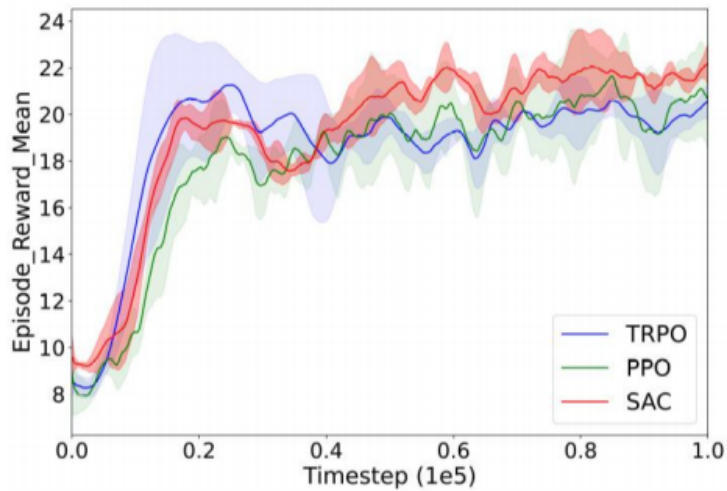

(a) Multi-lane highway

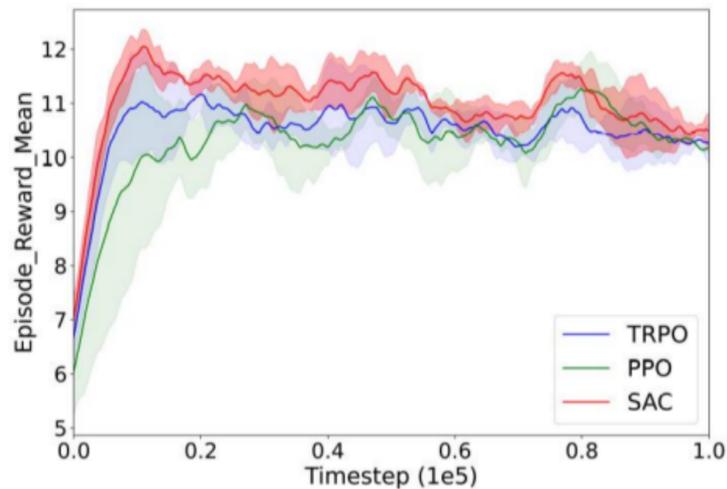

(b) Unsignalized T-intersection

Fig. 3: Reward curves of RL algorithms during training

TABLE II: Comparison of total number of collisions

| Scenario | Number of SVs | BIDA | MCTS | SAC | MOBIL/IDM |
|---|---|---|---|---|---|
| Multi-lane highway | 10 | **0** | 2 | 12 | 4 |
| | 20 | **1** | 8 | 21 | 10 |
| | 30 | **5** | 18 | 28 | 16 |
| | Average | **2** | 9.33 | 20.33 | 10 |
| Unsignalized T-intersection | 6 | **1** | 9 | 17 | 21 |
| | 8 | **3** | 14 | 20 | 26 |
| | 10 | **4** | 18 | 24 | 29 |
| | Average | **2.67** | 13.67 | 20.33 | 25.33 |

TABLE III: Comparison of total number of lane-changing/intersection-passing time (seconds)

| Scenario | Number of SVs | BIDA | MCTS | SAC | MOBIL/IDM |
|---|---|---|---|---|---|
| Multi-lane highway | 10 | **6.06** | 6.38 | 6.15 | 6.85 |
| | 20 | **6.29** | 6.89 | 6.39 | 6.72 |
| | 30 | **6.47** | 6.73 | 6.63 | 6.60 |
| | Average | **6.27** | 6.67 | 6.39 | 6.72 |
| Unsignalized T-intersection | 6 | **6.79** | 8.05 | 7.12 | 7.39 |
| | 8 | **9.10** | 9.82 | 9.43 | 9.27 |
| | 10 | **11.82** | 12.85 | 12.15 | 12.45 |
| | Average | **9.24** | 10.24 | 9.57 | 9.70 |

From Figure 3, it is evident that all three algorithms achieve convergence. Among them, SAC, which employs an entropy-maximizing RL method, exhibits superior performance by exploring the policy space more extensively in search of a global optimum, resulting in higher episodic rewards compared to PPO and TRPO. Therefore, we integrate the policy and value networks trained via SAC with the interactive search tree for decision-making inference in this paper.

C. Experimental Results

In this subsection, we compare the performance of the proposed BIDA method with various baseline algorithms, including the classical MCTS method, the RL method trained via SAC, the rule-based minimizing overall braking impact lane-change (MOBIL) method [25] in the multi-lane highway scenario, and the rule-based IDM method in the unsignalized T-intersection scenario. For each scenario, 200 test episodes are conducted per algorithm under different traffic densities, resulting in a total of 1600 test episodes. The experimental results are evaluated based on three key metrics: safety, efficiency, and interaction rationality.

1) Safety Analysis: The number of collisions occurring during test episodes is used as the safety metric. As can be seen in Table II, the classical MCTS method achieves lower collision rates than SAC and MOBIL by simulating multi-step future scenarios. However, due to the lack of a policy network to guide the exploration of more promising nodes, it results in a higher number of collisions than our BIDA. In contrast, our BIDA significantly reduces collisions by predicting dangerous scenarios and guiding policy generation accordingly. Compared to MCTS, our BIDA's collision rate is reduced by 78.57% in the multi-lane highway scenario and by 80.49% in the unsignalized T-intersection scenario, leading to safer decisions made by AVs.

2) Efficiency Analysis: In both scenarios, efficiency is measured by lane-changing time and intersection-passing time. The experimental results are illustrated in Table III. It can be observed that the rule-based methods (i.e., MOBIL and IDM) lack self-learning and optimization capabilities, leading to overly conservative driving decisions. MCTS relies heavily on extensive simulations and tree search to predict future decisions, which results in higher computation complexity and lower efficiency. Our BIDA achieves the best efficiency, primarily due to its combination of RL for policy optimization and its predictive capability for high-risk scenarios. Overall, the BIDA improves lane-changing efficiency by 17.74% , and intersection-passing efficiency by 14.15% compared to the second-best SAC method, respectively.

3) Interaction Rationality Analysis: In this study, we define an "invasive action" as an instance where the ego vehicle causes another interactive vehicle to apply emergency braking ($a < -2\,\text{m/s}^2$). The total number of invasive actions across all test episodes is recorded in Table IV. Our BIDA demonstrates superior interaction rationality by leveraging RL's trial-and-error exploration and its predictive risk mechanism to anticipate potential hazards. Invasive actions are penalized during training, allowing the BIDA to interact more smoothly with other vehicles and improve overall traffic stability and safety, while maintaining traffic efficiency.

TABLE IV: Comparison of total number of invasive actions

| Scenario | Number of SVs | BIDA | MCTS | SAC | MOBIL/IDM |
|---|---|---|---|---|---|
| Multi-lane highway | 10 | 0 | 0 | 2 | 1 |
| | 20 | 0 | 3 | 6 | 1 |
| | 30 | 0 | 7 | 9 | 4 |
| | Average | 0 | 3.33 | 5.67 | 2 |
| Unsignalized T-intersection | 6 | 0 | 3 | 5 | 9 |
| | 8 | 0 | 4 | 8 | 12 |
| | 10 | 0 | 9 | 11 | 15 |
| | Average | 0 | 5.33 | 8 | 12 |

## IV. CONCLUSION

In this paper, we propose a bi-level interaction decision-making algorithm (BIDA) integrating interactive search tree with deep reinforcement learning to optimize decision-making in dynamic traffic scenarios. During the closed-loop simulation in CARLA, a rolling optimization process is conducted. Experiments show that our BIDA enhances interaction rationality, efficiency, and safety in multi-lane highway and unsignalized T-intersection. However, it should be noted that our BIDA still experiences a small number of collisions under high-density conditions. Future work will focus on improving BIDA's robustness in diverse traffic environments and exploring its integration with vehicle-to-vehicle communication and multi-agent reinforcement learning for real-time cooperation in large-scale networks.

## ACKNOWLEDGMENTS

This work is supported in part by the National Nature Science Foundation of China (No. 62273256) and the Fundamental Research Funds for the Central Universities.